\documentclass[runningheads,a4paper]{llncs}

\usepackage{amssymb}
\usepackage{graphicx}

\usepackage{url}
\urldef{\mail}\path|tim@tim-taylor.com|

\usepackage{natbib}
\makeatletter
\renewcommand\bibsection%
{
  \section*{\refname
    \@mkboth{\MakeUppercase{\refname}}{\MakeUppercase{\refname}}}
}
\makeatother

\begin{document}

\mainmatter  

\title{Requirements for Open-Ended Evolution in Natural and Artificial
Systems}

\titlerunning{Requirements for Open-Ended Evolution}

\author{Tim Taylor}
\authorrunning{Tim Taylor}

\institute{Department of Computer Science and York Centre for Complex
  Systems Analysis\\ University of York, United Kingdom\\
\mail}

\maketitle

\begin{abstract}

Open-ended evolutionary dynamics remains an elusive goal for
artificial evolutionary systems. Many ideas exist in the biological
literature beyond the basic Darwinian requirements of variation,
differential reproduction and inheritance. I argue that these ideas
can be seen as aspects of five fundamental requirements for open-ended
evolution: (1) robustly reproductive individuals, (2) a medium
allowing the possible existence of a practically unlimited diversity
of individuals and interactions, (3) individuals capable of producing
more complex offspring, (4) mutational pathways to other viable
individuals, and (5) drive for continued evolution. I briefly discuss
implications of this view for the design of artificial systems with
greater evolutionary potential.

\end{abstract}

\section{Introduction}
If there is one lesson to be learned from the first 60 years of
research into the evolution of digital organisms, it is that the
classic Darwinian ingredients of \emph{variation}, \emph{differential
  reproduction} and \emph{inheritance} are not, in themselves,
sufficient for producing open-ended dynamics in which new, surprising,
and sometimes more complex organisms continue to appear
\citep{Taylor:DigitalGenesis}.\footnote{In this paper I will use an
  informal definition of open-ended evolution as ``evolutionary
  dynamics in which new, surprising, and sometimes more complex
  organisms continue to appear.''} 

Most evolutionary artificial life systems tend to rather quickly
reach a quasi-stable state beyond which no qualitatively new
innovations are seen to appear \citep{Taylor:EvolVirtWorlds}. None has
displayed dynamics which might be regarded as the holy grail of
artificial life, where the continued evolution of novel forms is so
interesting that the researcher is unwilling to press the ``off'' switch.

Various artificial life researchers have started to look at different
aspects of the biological world for the missing ingredients. At the
same time, our understanding of processes important in biological
evolution has been greatly supplemented by new research in many areas,
including epigenetics \citep{Jablonka:Evolution}, non-coding regions of DNA
\citep{Comfort:Genetics}, neutral evolutionary networks \citep{Wagner:Origins},
facilitated variation \citep{Gerhart:Theory}, niche construction
\citep{OdlingSmee:Niche}, and others. 

While these new research directions are exciting and promise new
insights into the important ingredients of biological evolution, the
underlying simplicity of the Darwinian picture of variation, differential
reproduction and inheritance soon disappears in the panoply of new
ideas. Of course, that might just be the price we have to pay for a
deeper understanding of evolution---biology, unlike the physical
sciences, is an historically contingent subject that can be fiercely
resistant to Occam's razor. On the other hand, it may  be that these
new ideas are all jigsaw pieces of a still simple, if somewhat expanded,
framework in which we can understand biological evolution.

In the following section, I suggest that there are five fundamental
requirements for a system to exhibit open-ended evolution. I show how
the various ideas mentioned above fit into this picture, discuss
how they relate to past work in artificial life, and suggest various
directions that are indicated for future research.

\section{Requirements}

At a very general level, the following five
features are necessary, and I claim sufficient, for a system to
exhibit open-ended evolutionary dynamics:\footnote{This list is a
  refinement of the ideas presented in \citep{Taylor:Exploring}.}

\begin{itemize}
\item Robustly reproductive individuals.
\item A medium allowing the possible existence of a practically
  unlimited diversity of individuals and interactions, at various
  levels of complexity.
\item Individuals capable of producing more complex offspring.
\item An evolutionary search space which typically offers mutational
  pathways from one viable individual to other viable (and potentially
  fitter) individuals.
\item Drive for continued evolution.
\end{itemize}

Each of these features is discussed below.


\subsection{Robustly reproductive individuals}

The basic components of any evolutionary system are individual entities
that can catalyse the production of (sometimes imperfect) copies of
themselves. Successful individuals must be robust enough to survive in
their environment until they have performed at least one
reproduction. In order for an evolutionary process to be sustained,
there must be at least some such robustly reproducing individuals in
the population.\footnote{Note that this requirement relates to the
  robustness of an \emph{individual} to survive in its (potentially variable)
  environment. A separate consideration is the robustness of a
  \emph{population} of individuals to cope with changing environments
  over evolutionary timescales; such population robustness is
  addressed in Section~\ref{sct-mutational}.}

While this may appear to be a fairly basic statement, the question of
what are the appropriate ways to achieve robustness in artificial life
systems has not often received the attention it deserves. Von
Neumann's self-reproducing automata \citep{VonNeumann:Theory}, and
other systems of self-reproduction in 2D cellular automata, are
generally not robust: they do not engage in self-maintenance and
self-repair, and are susceptible to disruptive perturbations from
neighbouring individuals. Hence, while these systems might possess
some desirable theoretical evolutionary capacity (see
Section~\ref{sct-more-complex-offspring}), in practice they are
evolutionary non-starters. 

Digital organism systems such as Tierra and Avida hard-wire robustness
into the system by not granting individuals write-access to other
parts of memory (except in the special case where some new memory has
been allocated for reproduction). This was a critical design decision
that allowed prolonged evolution to happen in these systems, in
contrast to predecessors such as Core War, where individuals could
overwrite each other with no such restrictions
\citep{Ray:Approach}. However, by hard-wiring write protection into
the system, programs in Tierra and Avida become relatively isolated
from each other, with consequences for what kinds of interactions are
possible.\footnote{This therefore also has consequences for the
 degree of drive for continued evolution (see Section~\ref{sct-drive}).} 

Biological organisms need to actively maintain their organisation
against the disruptive pull of the second law of
thermodynamics. Concerns of entropy increase are not immediately
applicable to digital organisms, unless entropy is intentionally built
into the digital physics of the system.\footnote{Although note that in
  some physically-inspired models of computation such as
  \emph{conservative logic}, there are more clearly defined analogies
  of heat dissipation and entropy \citep{Fredkin:Conservative}.}
If entropy \emph{was} built into an artificial life system, it would mean
that the digital organisms would have to concern themselves with
self-maintenance, and that most structures would naturally decay
without the need for arbitrary mechanisms like reaper queues. This
would entail the organisation of digital organisms more closely
corresponding to the characterisation of living organisation as
self-building, self-maintaining and self-reproducing systems, e.g.\ 
\citep{Varela:Autopoiesis, Rosen:Life, Ganti:Principles}. 

An significant open question for artificial life research is
understanding the importance of topics such as entropy and
self-maintenance for open-ended evolution.

\subsection{A medium allowing the possible existence of a practically
  unlimited diversity of individuals and interactions, at various
  levels of complexity}
\label{sct-medium}

A clear requirement for open-ended evolution is that many different
types of organism must be conceivable within the system. The medium in
which the evolutionary process is unfolding must allow the possibility
of a practically unlimited diversity of organism organisations,
processes and interactions. 

Much previous work within artificial life has concentrated on the
ability of organisms to evolve complex computational and information
processing capabilities, such as the ability of digital organisms in
Avida to solve logic functions \citep{Lenski:Evolutionary} or the
evolution of complex neural network-driven behaviour in systems such
as Polyworld \citep{Yaeger:Evolutionary} and Geb
\citep{Channon:Unbounded}. 

However, it is restrictive to only consider the evolution of
information processing capabilities. Some of the most remarkable
events in biological evolutionary history have involved the evolution
of new ways of interacting with the environment via new sensors and
effectors. The geochemical-physical medium in which biological
evolution unfolds offers an enormously rich source of complex
dynamics, across many different modalities of phenomena, that may
potentially be exploited by organisms to promote their survival and
reproduction. 

The need for complex environments for the production of interesting
evolution in artificial life systems has been recognised right back to
the earliest work in the area. \cite{Barricelli:Numerical2} spoke in
terms of adding ``toy bricks'' to the environment to 
allow his digital organisms to evolve interesting behaviours.  

In addition, the major transitions in evolution identified by
\cite{MaynardSmith:MajorTransitions} involve changes in the organisation of
individuals over evolutionary time. Hence, open-ended artificial
life systems should allow the organisation of individual
organisms to evolve as well.

Many issues arise when designing complex virtual environments in which
organisms can evolve to access and exploit that complexity for their
own ends. These include questions such as whether the medium should
have ``messy'' processes with side effects, to allow for the
serendipity often apparent in biological evolution, and ``matter of
degree'' rather than ``all or nothing'' processes to allow for
gradual evolution---described by \cite{Dennett:Turing} as ``sorta''
evolution. Further issues concern the origin of signs and
signals, i.e.\ biosemiosis \citep{Hoffmeyer:Semiotic}, and the
representational relationship between organisms and environment such
that aspects such as new sensors and effectors can evolve without
being ``programmed in'' by the designer \citep{Taylor:Redrawing}.

\subsection{Individuals capable of producing more
  complex offspring} 
\label{sct-more-complex-offspring}

Beyond having a medium in which a wide variety of organism designs
could possibly exist, in order for complex adaptations to
\emph{evolve} from simple progenitors, it must be \emph{possible} for
an individual (or multiple individuals) to produce offspring that are
more complicated than their parent(s). 

There are (at least) two ways in which this may occur:

\begin{itemize}
\item A single individual is capable of producing an offspring of
  greater complexity than itself.
\item Two or more individuals are jointly capable of producing an
  offspring of greater complexity than any one of its parents.
\end{itemize}

The first solution is exactly the issue addressed by
\cite{VonNeumann:Theory} in his \emph{Theory of Self-Reproducing
  Automata}. The fundamental requirement identified by 
von Neumann is that the inherited information-bearing structures must
be involved in two distinct processes: (1) they are \emph{interpreted} by
the phenotype's machinery as instructions to guide the construction of
an individual, and (2) they are copied \emph{uninterpreted} from parent to
offspring.

Seen in this general light, we can say that von Neumann's requirements
are satisfied by biological cells (in 3D), by his proposed
self-reproducing cellular automata (in 2D), and by digital organisms
such as those in Tierra (in 1D). Note, however, that in the case of
Tierra, the interpretation machinery is hard-coded into an organism's
``virtual CPU'' is is therefore not evolvable. In addition, it is also
desirable to allow for the evolution of other aspects of the
evolutionary process itself, such as allowing new forms of genetic
transmission, evolution of the organisation of the genome, evolution
of mutation rates, etc.\ \citep{Hindre:New}. Hence, issues of
explicit versus implicit encoded, embeddedness in the medium, etc.,
are also important concerns here \citep{Taylor:EvolVirtWorlds}.

Biological examples of the second solution include horizontal gene
transfer (HGT) and symbiogenesis. These processes are much less well
explored in the artificial life literature, despite their significance
in biological evolution and the fact that they provide a feasible
complementary (or alternative) route to increased complexity.

\subsection{Mutational pathways to other viable individuals}
\label{sct-mutational}

For an open-ended evolutionary process, it is insufficient for
individuals to have the theoretical capacity for producing more
complicated offspring. The fitness landscape of the system must be
such that there are often viable mutational pathways open to
individuals, leading to different individuals that are of roughly the
same fitness, or of higher fitness, than their parents. That is, there
must often be the opportunity for adaptive, or at least neutral,
evolution. Otherwise, the evolutionary process will often get stuck in 
local optima (dead ends) beyond which no further change is possible.

While this has been understood for a long time---e.g.\
\cite{Rensch:Evolution} discussed the need for ``improvements allowing
further improvement''---the task of understanding the requirements
for a fitness landscape to have this property is now a very active
area of research.

A wide variety of work can be seen as contributing to this topic,
including \cite{Wagner:Origins}'s work on evolutionary innovations and
neutral networks, a wide range of work on the evolution of evolution,
e.g.\ \citep{Hindre:New}, evolvable genotype--phenotype mappings,
e.g.\ \citep{Gerhart:Theory, Wagner:Complex, Wills:Genetic}, and
major transitions, e.g.\ \citep{MaynardSmith:MajorTransitions}. Also
relevant is work on understanding how complex structures can evolve from simpler
components in modular, hierarchical and nearly-decomposable systems, e.g.\
\citep{Simon:Architecture, Watson:Compositional, Calcott:Other}, and
related work on semiosis in the origin of modular and loosely coupled systems,
e.g.\ \citep{Auletta:TopDown}. \cite{Conrad:Geometry} has also argued
that redundant, loosely coupled systems can aid evolvability by creating
``extradimensional bypasses'' that prevent evolution from getting
stuck in local optima.


The importance of exaptation---where an existing phenotypic structure
becomes selected for a different function---is well recognised in biology
\citep{Gould:Exaptation, Whitacre:Degeneracy}. 
A challenge for achieving open-ended evolution in artificial systems
is to work with structures that potentially have multi-functional
properties, perhaps in different domains of interaction
\citep{Taylor:EvolVirtWorlds}.


All of the topics mentioned here (and many others too) provide us with
ideas of how to create artificial evolutionary systems in which
individuals have \emph{room to move} as they explore the evolutionary
landscape.


\subsection{Drive for continued evolution}
\label{sct-drive}

Even with the first four requirements in place, a continued drive is
required to force the system to explore new states.


To create \emph{any} drive in the system at all, selection
pressure must exist. In general, this can be brought about by
competition for some kind of limited resource (which may be matter,
energy, space), or through environmental conditions, etc. Selection
creates an adaptive landscape in which some variations of organism are
favoured over others.  

In order to achieve \emph{continued} drive, the individuals must 
experience a \emph{changing}
adaptive landscape \citep{Waddington:Paradigm}. In biological
populations this is brought about by other individuals being part of
the ecological environment---those individuals are also evolving, and
can alter the fitness landscape by direct interaction, 
e.g.\ co-evolution \citep{Thompson:Coevolutionary}, or indirectly
through their actions, e.g.\ ecosystem engineering
\citep{Jones:Positive} and niche construction
\citep{OdlingSmee:Niche}. Changes can also come about
through (passive or active) diffusion of species to new environments
(e.g.\ migration). 

A changing adaptive landscape also has bearing on the available
mutational pathways of the system (Section~\ref{sct-mutational}), as
it will have consequences for what set of mutational neighbours of an
individual are now viable.\footnote{Indeed, this process has been
  observed experimentally in studies of virus--bacteria coevolution
  \citep{Meyer:Repeatability,Thompson:Role}.}  

Some kinds of drive will push the system towards higher complexity
(e.g.\ co-evolutionary arms races), whereas others will lead to change
but not necessarily higher complexity. Whether the latter counts as
``open-ended evolution'' depends on one's definition.

Many artificial evolutionary systems lack the rich
\emph{connectedness} of individuals brought about by ecological
interactions, niche construction, etc., and this is no doubt part of
the explanation of why open-endedness remains elusive in those systems
\citep{Taylor:EvolVirtWorlds}.  In addition, if we wish to engineer 
artificial evolutionary systems aimed at solving particular problems,
an important question is how to appropriately introduce some kind of
extrinsic selection (e.g. fitness function), rather
than relying on purely intrinsic natural selection, while still
retaining an effective drive at each step of the process. 

\section{Conclusion}

While it is clear that the requirements for open-ended evolution
extend far beyond the basic Darwinian demands of variation,
differential reproduction and inheritance, I have argued that 
these additional ideas can be seen as aspects of five basic
requirements: (1) robustly reproductive individuals, (2) a medium
allowing the possible existence of a practically unlimited diversity
of individuals and interactions, (3) individuals capable of producing
more complex offspring, (4) mutational pathways to other viable
individuals, and (5) drive for continued evolution.

While advances in the evolutionary potential of
artificial systems can come about by careful consideration of the
details of all of the topics and theories discussed, it is useful to
consider these five basic features as the foundation upon which
open-ended evolution can be achieved. 

\footnotesize
\bibliographystyle{apalike}
\bibliography{taylor-evoevo-oee}

\end{document}